\documentclass[journal]{IEEEtran}

\usepackage{pgfplots}
\pgfplotsset{compat=1.18}
\usepackage{amsmath,amsfonts}
\usepackage{algorithmicx}
\usepackage{algorithm}
\usepackage{algpseudocode}

\usepackage{booktabs}

\usepackage{array}
\usepackage[caption=false,font=scriptsize,labelfont=sf,textfont=sf]{subfig}
\usepackage{textcomp}
\usepackage{stfloats}
\usepackage{url}
\usepackage{verbatim}
\usepackage{graphicx}
\usepackage{cite}
\usepackage[utf8]{inputenc}
\usepackage{ragged2e}
\usepackage{microtype}
\usepackage{pgfplots}
\usepackage{comment}
\usepackage{xcolor}
\usepackage{colortbl}
\definecolor{gray}{rgb}{0.9,0.9,0.9}

\usepackage{adjustbox}
\usepackage{amssymb}
\usepackage{multirow}

\begin{document}

\title{MR-Occ: Efficient Camera-LiDAR 3D Semantic Occupancy Prediction Using Hierarchical Multi-Resolution Voxel Representation}

\author{Minjae Seong$^{1*}$, Jisong Kim$^{2*}$, Geonho Bang$^{1*}$, Hawook Jeong$^3$,  and Jun Won Choi$^4$ \\
\thanks{$^1$M. Seong and G. Bang are with the Department of Artificial Intelligence, Hanyang University, 04753 Seoul, Republic of Korea. (e-mail: mjseong@spa.hanyang.ac.kr, ghbang@spa.hanyang.ac.kr)}

\thanks{$^2$J. Kim is with the Department of Electrical Engineering, Hanyang University, 04753 Seoul, Republic of Korea. (e-mail: jskim@spa.hanyang.ac.kr)}

\thanks{$^3$H. Jeong is with the RideFlux Inc., 07217 Seoul, Republic of Korea. (e-mail: hawook@rideflux.com)}

\thanks{$^4$J. W. Choi is with the Department of Electrical and Computer Engineering, Seoul National University, Seoul, 08826, Korea. (e-mail: junwchoi@snu.ac.kr)\textit{(Corresponding author: Jun Won Choi)}}

\thanks{$^*$denotes equal contribution.}
}

\markboth{Journal of \LaTeX\ Class Files,~Vol.~14, No.~8, April~2024}%
{Shell \MakeLowercase{\textit{et al.}}: A Sample Article Using IEEEtran.cls for IEEE Journals}

\IEEEpubid{0000--0000/00\$00.00~\copyright~2024 IEEE}

\maketitle

\begin{abstract}
Accurate 3D perception is essential for understanding the environment in autonomous driving. Recent advancements in 3D semantic occupancy prediction have leveraged camera-LiDAR fusion to improve robustness and accuracy. However, current methods allocate computational resources uniformly across all voxels, leading to inefficiency, and they also fail to adequately address occlusions, resulting in reduced accuracy in challenging scenarios. We propose MR-Occ, a novel approach for camera-LiDAR fusion-based 3D semantic occupancy prediction, addressing these challenges through three key components: Hierarchical Voxel Feature Refinement (HVFR), Multi-scale Occupancy Decoder (MOD), and Pixel to Voxel Fusion Network (PVF-Net). HVFR improves performance by enhancing features for critical voxels, reducing computational cost. MOD introduces an `occluded' class to better handle regions obscured from sensor view, improving accuracy. PVF-Net leverages densified LiDAR features to effectively fuse camera and LiDAR data through a deformable attention mechanism. Extensive experiments demonstrate that MR-Occ achieves state-of-the-art performance on the nuScenes-Occupancy dataset, surpassing previous approaches by $\mathbf{+5.2\%}$ in IoU and $\mathbf{+5.3\%}$ in mIoU while using fewer parameters and FLOPs. Moreover, MR-Occ demonstrates superior performance on the SemanticKITTI dataset, further validating its effectiveness and generalizability across diverse 3D semantic occupancy benchmarks.
\end{abstract}


\section{Introduction}
\IEEEPARstart{3}{D} semantic occupancy prediction task aims to predict the occupancy and semantic information of fine-grained voxels surrounding the ego vehicle. This task provides a comprehensive volumetric scene representation that can be utilized by the path planner to enhance the safety of autonomous driving. While early research primarily focused on unimodal approaches using either LiDAR or camera data, recent methods have used multi-modal sensors, such as LiDAR and camera, to achieve more robust and accurate predictions.

The key challenge in camera-LiDAR fusion for 3D semantic occupancy prediction lies in constructing robust 3D representations that effectively handle the distinct characteristics of each modality. LiDAR provides accurate 3D positional information, enabling accurate occupancy prediction, while cameras offer rich semantic details for object classification. Previous studies \cite{openoccupancy, coocc, occgen} have focused on integrating these two modalities by using 3D voxel representations generated from LiDAR data. To fuse camera features with 3D LiDAR features, these approaches predict depth from images, transform them from 2D to 3D views to create voxel-based image features, and then align these features with the LiDAR data in the voxel space. Finally, the fused features are processed using 3D convolution-based modules.

\begin{figure}[t]
  \begin{center}
    \includegraphics[width=0.98\columnwidth]{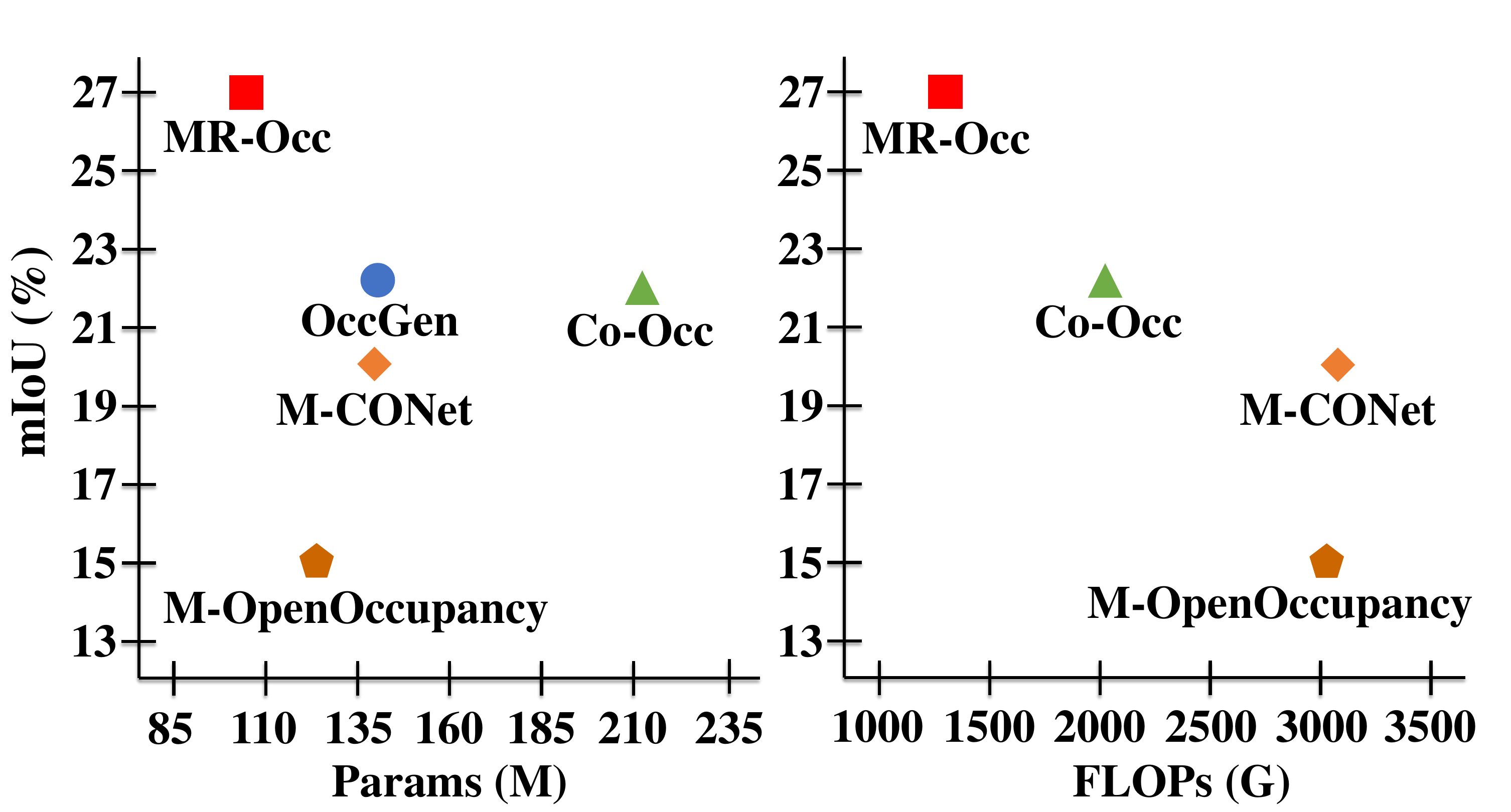}
  \end{center}
    \caption{{\bf Accuracy vs Efficiency (Params/FLOPs) on nuScenes-Occupancy validation set.} MR-Occ achieves state-of-the-art performance with less computational cost than previous methods.}
    \label{fig:intro}
\end{figure}

\IEEEpubidadjcol

While these existing approaches have yielded great results, they encounter two main challenges. First, computational resources are uniformly allocated across all voxel representations, which is inefficient. In outdoor scenarios, only a small fraction of voxels are non-empty—approximately 2.47\% according to the nuScenes-Occupancy dataset \cite{openoccupancy}. Therefore, considering the distribution of data across 3D voxels is essential for achieving more efficient occupancy state predictions.

Second, existing methods have underestimated the challenges posed by occlusion in real-world scenarios, leading to notable prediction inaccuracies in regions obscured from sensor view. These methods tend to either overlook hidden areas or assume equal visibility across all voxels. Thus, distinguishing between visible and non-visible areas is essential for achieving robust and accurate model predictions in complex environments.

In this paper, we present MR-Occ, a novel approach for camera-LiDAR fusion-based semantic occupancy prediction. The key ideas of MR-Occ are three-fold. 

First, we introduce Hierarchical Voxel Feature Refinement (HVFR) method, which selectively increases the resolution of voxels for efficient voxel encoding. Beginning with low resolution voxels, HVFR identifies core voxels, that captures important regions in a scene, based on occupancy confidence scores. By subdividing these critical voxels into smaller ones at finer resolutions, HVFR can capture important details and more refined voxel representation while maintaining low computational cost.

Second, we found that regions not visible to sensors due to occlusion by objects pose significant challenges for 3D occupancy prediction models. These models may need to rely on prior object knowledge or temporal information to determine occupancy in these ambiguous regions. When predictions for these regions are treated equally with those for non-ambiguous areas, the models may struggle to accurately predict occupancy. To address this issue, we divide Non-empty class into Occluded and Non-occluded classes, where Occluded class is assigned to voxels that are not visible to sensors but are labeled as Non-empty. By incorporating the Occluded class, we can train the model to classify a new class while treating the Occluded class as Non-empty class in the inference phase. This approach effectively reduces the burden on the models and improves occupancy prediction accuracy. Importantly, it does not require additional computational resources, as only a simple logic for labeling the ambiguity state is needed. 

Finally, we introduce the Pixel to Voxel Fusion Network (PVF-Net), an efficient feature fusion strategy that associates camera features with each voxel using a deformable attention mechanism guided by voxel queries generated from densified LiDAR features. Unlike most previous methods, which suffer from misalignment between camera and LiDAR features due to inaccurate depth estimation, our approach does not rely on depth prediction for feature fusion. This effectively resolves the misalignment issue, leading to significant performance gains.

MR-Occ achieves state-of-the-art performance on the nuScenes-Occupancy dataset for camera-LiDAR-based 3D semantic occupancy prediction task. It demonstrates remarkable performance gains of  \(\mathbf{+5.2\%}\) in IoU and \(\mathbf{+5.3\%}\) in mIoU over the previous best method, OccGen \cite{occgen} while requiring fewer parameters and FLOPs, as shown in Figure \ref{fig:intro}.

The key contributions of our work are as follows:
\begin{itemize}
    \item We propose MR-Occ, a new 3D semantic occupancy prediction model that effectively fuses LiDAR and camera features.
    \item We propose a Hierarchical Voxel Feature Refinement method that successively increases the resolution of core voxels that capture key scene information. Our approach significantly enhances the efficiency of voxel encoding.
    \item We propose a Multi-scale Occupancy Decoder that introduces a new Occluded state for voxels in the occupancy prediction problem. Our model is assigned a task of classifying Occluded state along with existing occupancy classes. This approach enables the model can effectively distinguish between occluded and non-occluded areas, thereby improving overall occupancy prediction accuracy.
    \item We propose a Pixel to Voxel Fusion Network that aligns 2D camera features with densified LiDAR features and fuses them with adaptive weights using deformable cross-attention.
    \item The code will be publicly available. 
\end{itemize}

\begin{figure*}[t]
    \centering
        \includegraphics[width=1.0\linewidth]{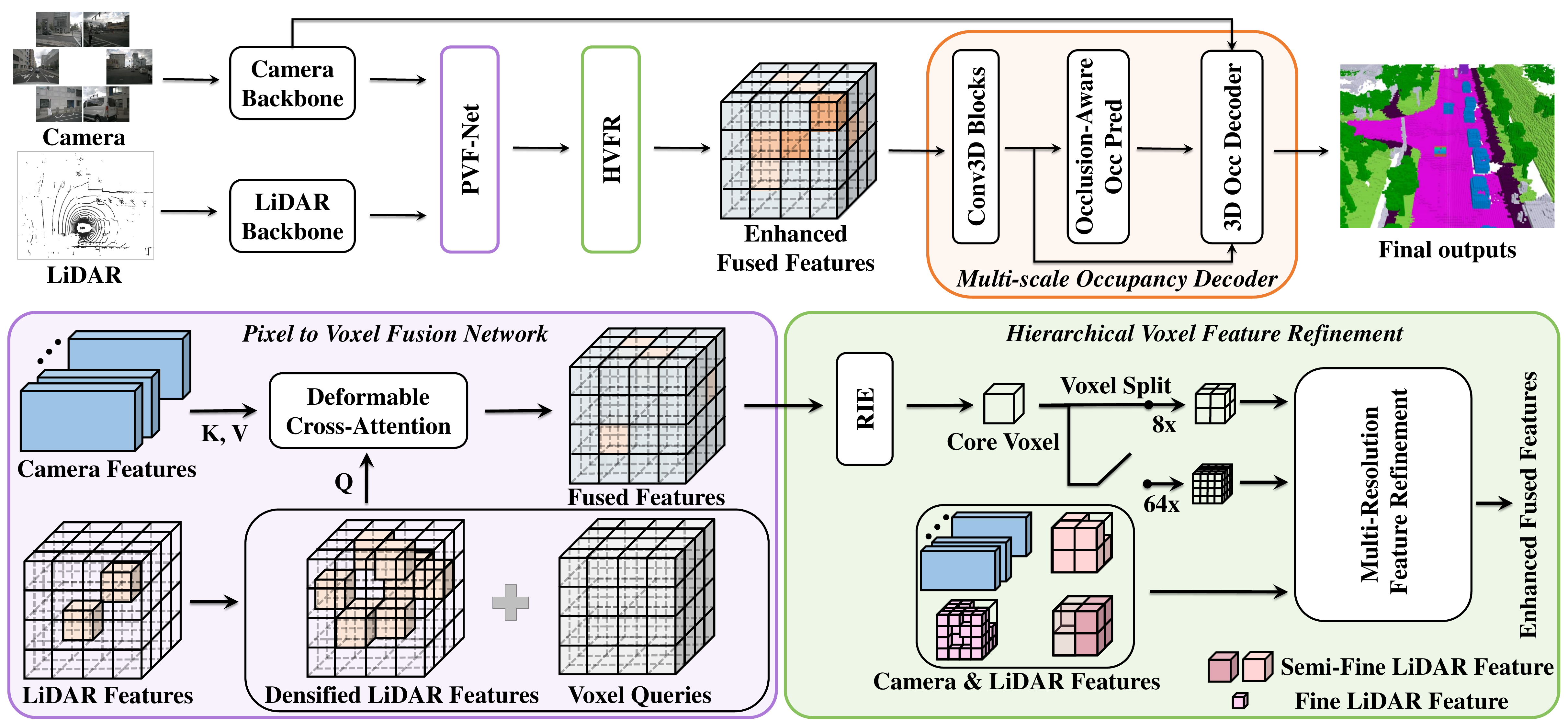}
        \caption{Overall architecture of MR-Occ: Camera and LiDAR features are extracted from modality-specific backbone networks. The PVF-Net densifies LiDAR features and adaptively fuses them with image features using a deformable cross-attention mechanism. The HVFR module uses Resolution Importance Estimator (RIE) to identify core voxels, and then enhances the fused features through Multi-Resolution Feature Refinement using these core voxels. Finally, Multi-scale Occupancy Decoder (MOD) predicts an `occluded' class for occluded areas and performs fine-grained occupancy prediction.}
    \label{fig:overall}
\end{figure*}

\section{Related Work}

\subsection{Unimodal 3D Semantic Occupancy Prediction}

3D Semantic Occupancy Prediction (SOP) has emerged as a pivotal task in autonomous driving, leading to diverse research on LiDAR and camera-based approaches. 
Early research in 3D SOP was predominantly focused on LiDAR-centric approaches \cite{lmscnet, s3cnet, js3cnet} based on the SemanticKITTI dataset \cite{semantickitti}. 
Recent advancements have expanded these approaches by transferring rich semantic information from multi-frame LiDAR models \cite{scpnet}, leveraging generative models \cite{occgen}, and constructing 2D tri-perspective view representations \cite{pointocc}.

In parallel, camera-based methods have garnered attention due to their cost-effectiveness and capacity to provide detailed visual information. A notable example is MonoScene \cite{monoscene}, which pioneered effective 3D SOP using only a single RGB image, thereby opening new research avenues. Recent camera-based approaches are primarily categorized into two paradigms: forward projection and backward projection. Forward projection explicitly maps 2D image features into 3D space by estimating depth and applying geometric transformations. OccFormer \cite{occformer} advances this approach by introducing a dual-path transformer that captures long-range dependencies and effectively handles dynamic 3D features. To optimize computational efficiency, methods such as \cite{flashocc, fastocc, sparseocc} combine 2D convolutions with channel-to-height transformations and utilize sparse 3D representations. In contrast, backward projection implicitly learns complex relationships between 2D and 3D representations, transforming 2D image features into 3D volumetric features through transformer architectures. Within this framework, TPVFormer \cite{tpvformer} and VoxFormer \cite{voxformer} employ tri-perspective views and depth-estimated voxel queries, respectively, to mitigate the computational cost associated with attention mechanisms.

\subsection{Multimodal 3D Semantic Occupancy Prediction}
While unimodal approaches have progressed significantly in 3D SOP, integrating multi-sensor data is crucial for comprehensive environmental understanding. Extensive research in 3D object detection, a closely related task to 3D SOP, has explored various sensor fusion strategies. Some approaches, inspired by multi-view 3D object detection, employ explicit view transformations \cite{lss} to align camera and LiDAR data within a unified Bird's Eye View (BEV) \cite{bevfusion, uvtr}. Others utilize attention-based methods \cite{transfusion, bevguide} with transformer architectures that use object and BEV queries to dynamically focus on relevant features from multiple sensor modalities.

While various sensor fusion techniques have been explored in 3D object detection task, multi-modal 3D SOP primarily utilizes explicit view transform methods to convert camera data into 3D voxel features. These camera 3D voxel features are then fused with LiDAR features in the 3D voxel space. CONet \cite{openoccupancy} employs adaptive fusion to integrate LiDAR and camera information, coupled with a coarse-to-fine prediction strategy. Similarly, Co-Occ \cite{coocc} introduces a Geometric- and Semantic-aware Fusion (GSFusion) module, enhancing LiDAR features by incorporating neighboring camera features through a K-nearest neighbors (KNN) search. These enriched LiDAR features are then concatenated with the original LiDAR and camera voxel features. In contrast, OccGen \cite{occgen} tackles the misalignment between camera and LiDAR features by applying a hard 2D-to-3D view transformation and utilizing a geometry mask.

Despite these advancements, existing multimodal approaches generate 3D voxel features separately for each modality, increasing computational costs during the fusion process. Additionally, by assigning equal importance to all voxels, these methods lead to unnecessary computations and inefficient use of resources, ultimately limiting performance improvements.

\section{MR-Occ}
The overall architecture of MR-Occ is illustrated in Figure 2. Our model is a camera-LiDAR fusion framework for 3D Semantic Occupancy Prediction that identifies core voxels and intensively enhances their features. In Section 3.1, we introduce the Pixel to Voxel Fusion Network (PVF-Net) method to fuse LiDAR voxel features with multi-camera features. Then, we present the Hierarchical Voxel Feature Refinement (HVFR) module for identifying core voxels and enhancing their features in Section 3.2. Finally, the Multi-scale Occupancy Decoder (MOD) for performing multi-scale 3D semantic occupancy prediction is described in Section 3.3.

\subsection{Pixel to Voxel Fusion Network}
Prior to integrating LiDAR and image features, we extract both features using separate backbone networks. For LiDAR data, we employ 3D sparse convolutional layers to compute multi-resolution voxel features $F_L = \{F_L^1, F_L^2, F_L^4\}$, where $F_L^i \in \mathbb{R}^{C_L^i \times H_L^i \times W_L^i \times D_L^i}$. Here, $i$ represents the downsampling scale, $(H_L^i, W_L^i, D_L^i)$ correspond to the 3D spatial dimensions, and $C_L^i$ represents the number of channels. For camera data, we utilize a ResNet-50 backbone integrated with a Feature Pyramid Network (FPN) to derive multi-view features \( F_I \in \mathbb{R}^{N \times H_I \times W_I \times C_I} \), where \( N \) is the number of cameras, \( (H_I, W_I)\) are the 2D spatial dimensions of the feature maps, and \( C_I \) denotes the channel dimension. 

Existing camera-LiDAR methods \cite{occgen, coocc, openoccupancy} transform image features $F_I$ to 3D voxel representation to fuse with LiDAR features. 
However, this method can lead to positional misalignment as the 3D features derived from camera are inherently inaccurate. 
To address this issue, we introduce the Pixel-to-Voxel Fusion Network (PVF-Net), which enhances multi-modal fusion through densified LiDAR features. PVF-Net expands the receptive field around non-empty voxels, enabling LiDAR features to guide the seamless integration of 2D image features into 3D voxel representation.

First, we downsample LiDAR feature $F_L^4$ using 3D sparse convolutions to obtain $F_L^8$ and $F_L^{16}$. Let the non-empty voxel features of \( F_L^i \) be denoted as \( V_L^i \in \mathbb{R}^{C_L^i \times N^i} \), where \( N^i \) represents the number of non-empty voxels, and their corresponding indices as \( G^i \in \mathbb{Z}^{3 \times N^i} \) at each scale \( i \). We then concatenate these features across different scales
\begin{equation}
    \tilde{V}_L^4 = [V_L^4, V_L^8, V_L^{16}], \quad \tilde{G}_L^4 = [G^4, 2 \cdot G^8, 4 \cdot G^{16}],
\end{equation}
where \([ \cdot ]\) denotes the concatenation operation. Note that the factors 2 and 4 in $2 \cdot G^8$ and $4 \cdot G^{16}$ align the scales of $G^8$ and $G^{16}$ with $G^4$.
The final dense feature map $\tilde{F}_L^4$ is computed by averaging the overlapping non-empty features at each voxel location
\begin{equation}
    \tilde{F}_L^4(x,y,z) = \frac{1}{|S(x,y,z)|} \sum_{k \in S(x,y,z)} \tilde{V}_L^4(k),
\end{equation}
where \( S(x,y,z) \) denotes the set of indices $k$ for which the corresponding voxel in $\tilde{G}_L^4$ maps to the voxel \( (x,y,z) \).

Next, we fuse \( \tilde{F}_L^4 \) with the image features \( F_I \) using deformable cross-attention. In this process, 3D voxel queries are projected onto the image plane, where nearby pixel features serve as keys and values. To guide the alignment of 2D image feature within the 3D voxel space, we update the random queries \( Q_v \) by adding the densified LiDAR features \( \tilde{F}_L^4 \). The LiDAR guided query \( Q'_v \) is then formulated as
\begin{equation}
    Q'_v = \tilde{F}_L^4 + Q_v.
\end{equation}
The Pixel-to-Voxel fusion is then performed as
\begin{equation}
    F_M^4 = \frac{1}{|V_{\text{hit}}|} \sum_{i \in V_{\text{hit}}} \sum_{j=1}^{N_{\text{ref}}} \text{DA}(Q'_v(p), P(p, i, j), F_I^i),
\end{equation}
where \( \text{DA} \) represents the deformable attention function, \( Q'_v(p) \) denotes the LiDAR guided query at position \( p \), and \( P(p,i,j) \) projects it to the \( j \)-th reference point on the \( i \)-th camera view. \( F_I^i \) represents features from the \( i \)-th camera, \( N_{\text{ref}} \) is the number of reference points per query, and \( V_{\text{hit}} \) is the set of cameras where the projected point is visible.

\begin{figure}
    \centering
        \includegraphics[width=0.99\linewidth]{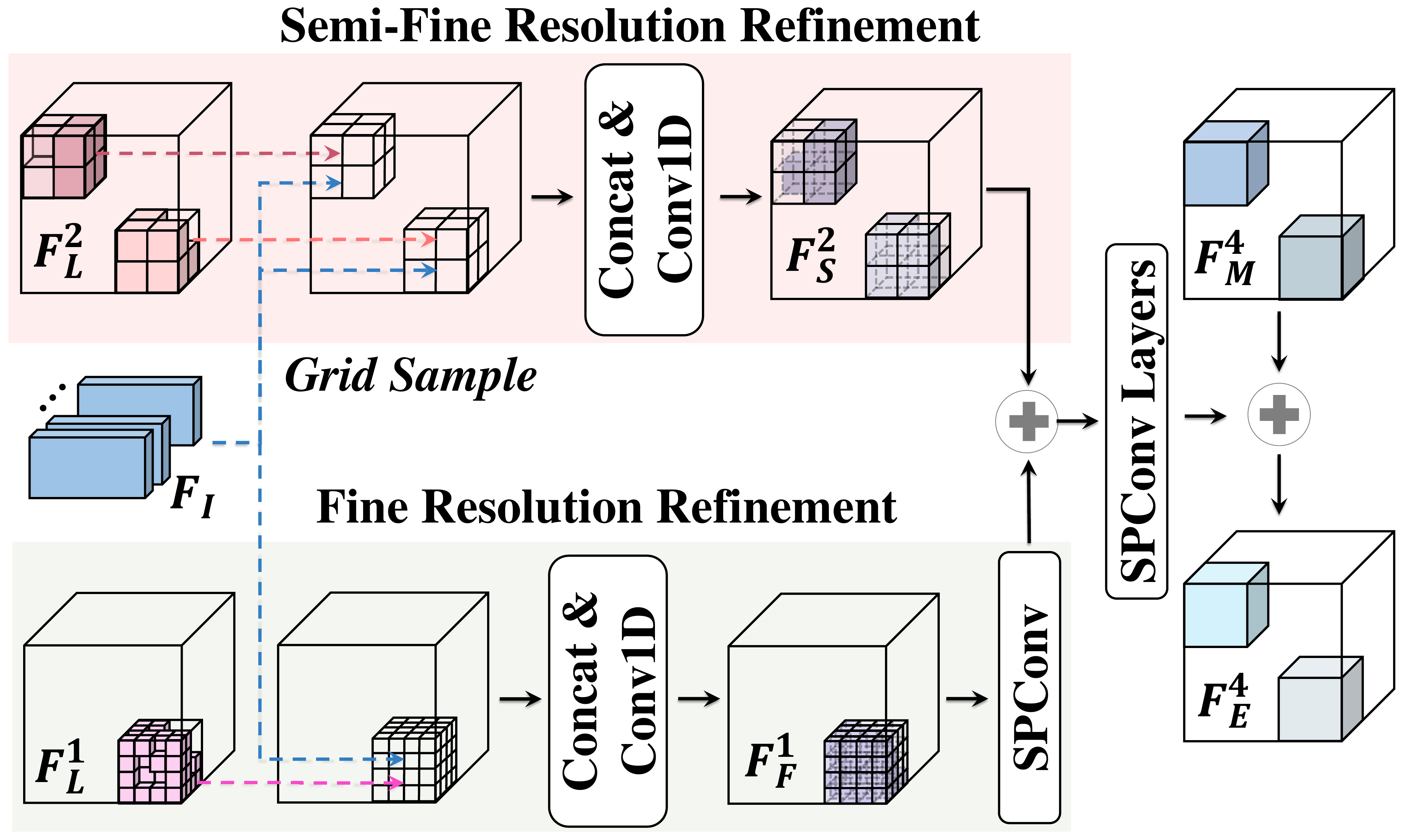}
        \caption{Multi-Resolution Feature Refinement module. The subdivided core voxels combine features sampled from the same resolution LiDAR features and camera features to capture fine-grained details. The multi-resolution features are fused based on a 3D sparse convolution.}
    \label{fig:feat_comp}
\end{figure}

\subsection{Hierarchical Voxel Feature Refinement}
Existing approaches often rely on downsampled fused features that are coarser than the ground truth voxel resolution, which hinders the accurate prediction of small objects and fine boundary details. While using finer-scale features could mitigate these issues, it would come at the cost of significantly increased computational complexity. To address this, we introduce a Hierarchical Voxel Feature Refinement (HVFR) module that adaptively refines the critical voxels within the feature map.

We first apply a Resolution Importance Estimator (RIE) to the fused feature map to determine the necessary level of detail for each voxel. This process yields a voxel-wise importance map \( R \) as given by the equation
\begin{equation}
    R = \sigma(\text{Conv}_{3D}(F_M^4)),
\end{equation}
where \( \sigma(\cdot) \) denotes the sigmoid function, and \( \text{Conv}_{3D}(\cdot) \) refers to a 3D convolutional layer. 
Voxels are selectively refined based on the value of $R(x, y, z)$  at each coordinate $(x, y, z)$ in comparison to the predefined thresholds $\tau_1$ and $\tau_2$. Specifically, if $R(x, y, z) \geq \tau_1$, the voxel is assigned to the Semi-fine Resolution set $\mathcal{S}$, and if $R(x, y, z) \geq \tau_2$, it is further included in the Fine Resolution set  $\mathcal{F}$.

For voxels in the Semi-fine Resolution set \( \mathcal{S} \), each voxel is uniformly subdivided into eight smaller sub-voxels. The corresponding LiDAR features are extracted from a finer resolution LiDAR feature map $ F_L^2 $, and the image features $ F_I^2 $ for these sub-voxels are obtained as follows
\begin{equation}
    F_I^2(x'_i, y'_i, z'_i) = \text{Proj}(F_I, (x'_i, y'_i, z'_i)), \quad i = 1, \dots, 8,
\end{equation}
where $\text{Proj}(\cdot)$ represents the operation of projecting the 3D positions $(x'_i, y'_i, z'_i)$ onto the image plane and retrieving the corresponding image features from $F_I$.
These two features are then concatenated channel-wise and passed through a $1 \times 1$ convolutional layer to generate a refined feature $F_S^2$ for the Semi-fine Resolution set, as follows
\begin{equation}
    F_{\mathcal{S}}^2 = \text{Conv}_{1\times 1}\left( [F_L^2, F_I^2] \right).
\end{equation}
For the Fine Resolution set $\mathcal{F}$, we apply a more detailed refinement process. Each voxel is uniformly subdivided into 64 finer sub-voxels, with the corresponding LiDAR features extracted from the finest resolution feature map $F_L^1$ and the image features obtained as follows
\begin{equation}
    F_I^1(x''_j, y''_j, z''_j) = \text{Proj}(F_I, (x''_j, y''_j, z''_j)), \quad j = 1, \dots, 64.
\end{equation}
The refined feature $F_F^1$ is then given by
\begin{equation}
    F_{\mathcal{F}}^1 = \text{Conv}_{1\times 1}\left( [F_L^1, F_I^1] \right).
\end{equation}
Finally, we apply multi-scale feature fusion to the hierarchical voxel features \( F_{\mathcal{S}}^2 \) and \( F_{\mathcal{F}}^1 \) to obtain selectively refined features for core voxels. This fusion process is formulated as follows
\begin{equation}
    F_E^4 = \text{SConv}_{2} \left(\text{SConv}_{1}\left(F_{\mathcal{F}}^1 \right) + F_{\mathcal{S}}^2 \right) + F_M^4,
\end{equation}  
where $\text{SConv}_{1}$ and $\text{SConv}_{2}$ represent 3D sparse convolutions applied sequentially to integrate the refined features with the original fused feature map, $F_M^4$.
The resulting feature map $F_E^4$ offers a more comprehensive scene representation, effectively synthesizing selectively refined details with broader contextual information.

\subsection{Multi-scale Occupancy Decoder}
3D semantic occupancy prediction requires dense predictions across the entire 3D space, including both visible and occluded voxels. However, previous studies have often overlooked visibility considerations in their occupancy state prediction frameworks. This oversight may limit the model's ability to fully understand the scene, potentially reducing prediction accuracy. To overcome this challenge, we introduce the Occlusion-aware Occupancy Prediction (OOP) module. This module classifies each voxel grid as empty, non-occluded, or occluded, thereby improving the model's robustness and overall performance.

We extend conventional voxel ground truth (GT) by integrating semantic classes with three additional labels: `non-occluded', `occluded', and `empty'. To assign these labels, we employ a ray-casting process with both LiDAR and camera data. Voxels containing LiDAR points or corresponding to projected image pixels are labeled `non-occluded' voxels, subsequent voxels that have already been assigned a class label are marked as `occluded'. The remaining voxels are labeled `empty'. The final label is determined by combining results from both modalities: a voxel is labeled `non-occluded' if either modality identifies it as such, `occluded' if both modalities agree, and `empty' if either modality classifies it as empty.

The enhanced fused features $F_{E}$ generated by the HVFR module are first processed through Conv3D blocks. These processed features are then fed into the Occlusion-Aware Occupancy Prediction module, which consists of a sequence of Conv3D-BN-ReLU-Conv3D layers. This module outputs $O^{4} \in \mathbb{R}^{D^4_{L} \times H^4_{L} \times W^4_{L} \times 21}$, where 18 of the 21 channels correspond to semantic occupancy classes, and the remaining 3 represent occlusion-aware classes (empty, non-occluded, or occluded). Finally, the 3D Occupancy Decoder, which is equivalent to the occupancy head in M-CONet, processes the voxels predicted as non-occluded or occluded in $O^{4}$. This decoder then predicts the final fine-grained semantic occupancy $O^{1} \in \mathbb{R}^{D^1_{L} \times H^1_{L} \times W^1_{L} \times 21}$, providing a detailed semantic classification for each relevant voxel in the scene.

\subsection{Loss Function}

Our model is optimized using a comprehensive loss function, following the method presented in CONet \cite{openoccupancy}.  The overall loss, \( \mathcal{L}_{\text{total}} \), is formulated as follows
\begin{equation}
    \mathcal{L}_{\text{total}} = \mathcal{L}_{\text{ce}} + \mathcal{L}_{\text{ls}} + \mathcal{L}_{\text{geo\_scal}} + \mathcal{L}_{\text{sem\_scal}} + \mathcal{L}_{\text{rs}} + \mathcal{L}_{\text{ocl}}
\end{equation}
where \( \mathcal{L}_{\text{ce}} \) represents the cross-entropy loss, and \( \mathcal{L}_{\text{ls}} \) denotes the Lovász-Softmax loss \cite{lovasz}, both of which are essential for semantic occupancy prediction. The affinity losses, \( \mathcal{L}_{\text{geo\_scal}} \) and \( \mathcal{L}_{\text{sem\_scal}} \) \cite{monoscene}, are incorporated to enhance scene-wise and class-wise metrics. Additionally, \( \mathcal{L}_{\text{rs}} \) is the binary cross-entropy loss employed for the resolution importance estimator, which is crucial for identifying key voxels. The ground truth for the resolution importance estimator is determined by assigning binary labels to each voxel, indicating whether it is occupied. Finally, \( \mathcal{L}_{\text{ocl}} \) is an occlusion-aware loss based on cross-entropy, designed for the Occlusion-aware Occupancy Prediction module to address visibility constraints from input sensors. Each term in the loss function contributes to improving the overall performance of the model in 3D semantic occupancy prediction.

\section{Experiments}

\subsection{Datasets}
We evaluate MR-Occ on two large-scale datasets: nuScenes-Occupancy \cite{openoccupancy} and SemanticKITTI \cite{semantickitti}. The nuScenes-Occupancy dataset provides dense semantic occupancy annotations for 1,000 scenes, with voxel grid annotations spanning \([-51.2\,\mathrm{m}, 51.2\,\mathrm{m}]\) in both X and Y directions, and \([-5\,\mathrm{m}, 3\,\mathrm{m}]\) in the Z direction, at a resolution of $512 \times 512 \times 40$. Each voxel is assigned one of 18 labels, comprising 17 semantic categories and 1 empty category. SemanticKITTI, based on the KITTI Odometry Benchmark \cite{kittiodometry}, consists of 22 sequences containing LiDAR scans and front camera images. The annotations use a $256 \times 256 \times 32$ voxel grid, with voxel grid coordinates spanning \([0\,\mathrm{m}, 51.2\,\mathrm{m}]\) along the X-axis, \([-25.6\,\mathrm{m}, 25.6\,\mathrm{m}]\) along the Y-axis, and \([-2\,\mathrm{m}, 4.4\,\mathrm{m}]\) along the Z-axis. Each voxel is assigned to one of 21 classes (19 semantic, 1 free, and 1 unknown).

We evaluate the performance of our method using two widely adopted metrics: 
\textit{Intersection over Union (IoU)} for geometric accuracy and \textit{mean Intersection over Union (mIoU)} for semantic-aware perception quality.

\paragraph{Intersection over Union (IoU)} 
IoU measures the voxel-level geometric accuracy as the ratio of the intersection volume to the union volume as
\begin{equation}
\text{IoU} = \frac{V_{\text{pred}} \cap V_{\text{gt}}}{V_{\text{pred}} \cup V_{\text{gt}}},
\end{equation}
where \( V_{\text{pred}} \) represents the predicted occupied voxels and \( V_{\text{gt}} \) represents the ground truth occupied voxels.

\paragraph{Mean Intersection over Union (mIoU)}
mIoU evaluates both occupancy prediction and its semantic consistency by averaging IoU across all \( C \) semantic classes as
\begin{equation}
\text{mIoU} = \frac{1}{C} \sum_{c=1}^{C} \frac{V_{\text{pred}, c} \cap V_{\text{gt}, c}}{V_{\text{pred}, c} \cup V_{\text{gt}, c}},
\end{equation}
where \( V_{\text{pred}, c} \) and \( V_{\text{gt}, c} \) denote the predicted and ground truth voxels for class \( c \), respectively.

\definecolor{lightyellow}{rgb}{1.0, 0.95, 0.6}  
\definecolor{lightgreen}{rgb}{0.7, 1.0, 0.7}  
\definecolor{cyan}{rgb}{0.0, 1.0, 1.0}  
\definecolor{pink}{rgb}{1.0, 0.75, 0.8}  
\definecolor{brown}{rgb}{0.6, 0.3, 0.1}  
\definecolor{orange}{rgb}{1.0, 0.6, 0.2}  
\definecolor{yellow}{rgb}{1.0, 1.0, 0.0}  
\definecolor{blue}{rgb}{0.0, 0.6, 1.0}  
\definecolor{olive}{rgb}{0.6, 0.6, 0.2}  
\definecolor{red}{rgb}{1.0, 0.0, 0.0}  
\definecolor{purple}{rgb}{0.5, 0.0, 0.5}  
\definecolor{magenta}{rgb}{1.0, 0.0, 1.0} 
\definecolor{gray}{rgb}{0.5, 0.5, 0.5}  
\definecolor{green}{rgb}{0.0, 0.6, 0.0} 
\definecolor{white}{rgb}{0.8, 0.8, 0.8}

\newcommand{\colorSquare}[1]{\tikz\draw[fill=#1,draw=none] (0,0) rectangle (0.8em,0.8em);}

\newcolumntype{C}{>{\centering\arraybackslash}p{2.3em}}

\newcolumntype{'}{!{\vrule width 0.2mm}}
\renewcommand{\arraystretch}{1.2}  

\setlength{\tabcolsep}{0.5mm} 

\begin{table*}[t]
\begin{center}
\begingroup
\arrayrulecolor{black} 
\setlength{\arrayrulewidth}{0.3mm} 

\fontsize{9pt}{10pt}\selectfont

\begin{tabular}{c ' c ' c c ' c c c c c c c c c c c c c c c c}
\toprule

Method & Mod. & IoU & mIoU & \rotatebox{90}{\colorSquare{orange}\hspace{0.3em}\textbf{barrier}} & \rotatebox{90}{\colorSquare{pink}\hspace{0.3em}\textbf{bicycle}} & \rotatebox{90}{\colorSquare{yellow}\hspace{0.3em}\textbf{bus}} & \rotatebox{90}{\colorSquare{blue}\hspace{0.3em}\textbf{car}} & \rotatebox{90}{\colorSquare{cyan}\hspace{0.3em}\textbf{const. veh.}} & \rotatebox{90}{\colorSquare{olive}\hspace{0.3em}\textbf{motorcycle}} & \rotatebox{90}{\colorSquare{red}\hspace{0.3em}\textbf{pedestrian}} & \rotatebox{90}{\colorSquare{lightyellow}\hspace{0.3em}\textbf{traffic cone}} & \rotatebox{90}{\colorSquare{brown}\hspace{0.3em}\textbf{trailer}} & \rotatebox{90}{\colorSquare{purple}\hspace{0.3em}\textbf{truck}} & \rotatebox{90}{\colorSquare{magenta}\hspace{0.3em}\textbf{drive. surf.}} & \rotatebox{90}{\colorSquare{gray}\hspace{0.3em}\textbf{other flat}} & \rotatebox{90}{\colorSquare{purple}\hspace{0.3em}\textbf{sidewalk}} & \rotatebox{90}{\colorSquare{lightgreen}\hspace{0.3em}\textbf{terrain}} & \rotatebox{90}{\colorSquare{white}\hspace{0.3em}\textbf{manmade}} & \rotatebox{90}{\colorSquare{green}\hspace{0.3em}\textbf{vegetation}}\\ 

\midrule
C-OpenOccupancy \cite{openoccupancy} & \multirow{5}{*}{C} & 19.3 & 10.3 & 9.9 & 6.8 & 11.2 & 11.5 & 6.3 & 8.4 & 8.6 & 4.3 & 4.2 & 9.9 & 22.0 & 15.8 & 14.1 & 13.5 & 7.3 & 10.2 \\
C-CONet \cite{openoccupancy} &  & 20.1 & 12.8 & 13.2 & 8.1 & 15.4 & 17.2 & 6.3 & 11.2 & 10.0 & 8.3 & 4.7 & 12.1 & 31.4 & 18.8 & 18.7 & 16.3 & 4.8 & 8.2 \\
SparseOcc \cite{sparseocc} &  & 21.8 & 14.1 & 16.1 & 9.3 & 15.1 & 18.6 & 7.3 & 9.4 & 11.2 & 9.4 & 7.2 & 13.0 & 31.8 & 21.7 & 20.7 & 18.8 & 6.1 & 10.6 \\
C-OccGen \cite{occgen} &  & 23.4 & 14.5 & 15.5 & 9.1 & 15.3 & 19.2 & 7.3 & 11.3 & 11.8 & 8.9 & 5.9 & 13.7 & 34.8 & 22.0 & 21.8 & 19.5 & 6.0 & 9.9 \\
\rowcolor{blue!10} 
C-MR-Occ & & 25.6 & 16.2 & 17.3 & 9.9 & 16.8 & 21.2 & 8.2 & 12.7 & 12.9 & 10.1 & 7.5 & 14.3 & 38.9 & 25.3 & 24.7 & 20.6 & 8.0 & 11.2 \\
\midrule
L-OpenOccupancy \cite{openoccupancy} & \multirow{4}{*}{L} & 30.8 & 11.7 & 12.2 & 4.2 & 11.0 & 12.2 & 8.3 & 4.4 & 8.7 & 4.0 & 8.4 & 10.3 & 23.5 & 16.0 & 14.9 & 15.7 & 15.0 & 17.9 \\
L-CONet \cite{openoccupancy}& & 30.9 & 15.8 & 17.5 & 5.2 & 13.3 & 18.1 & 7.8 & 5.4 & 9.6 & 5.6 & 13.2 & 13.6 & 34.9 & 21.5 & 22.4 & 21.7 & 19.2 & 23.5 \\
L-OccGen \cite{occgen}& & 31.6 & 16.8 & 18.8 & 5.1 & 14.8 & 19.6 & 7.0 & 7.7 & 11.5 & 6.7 & 13.9 & 14.6 & 36.4 & 22.1 & 22.8 & 22.3 & 20.6 & 24.5 \\
\rowcolor{blue!10} 
L-MR-Occ & & \textbf{35.7} & 24.1 & 28.6 & 13.6 & 22.1 & 29.0 & 13.5 & 20.4 & 26.4 & 16.1 & 18.3 & 23.2 & \textbf{38.5} & 25.1 & \textbf{26.2} & 25.7 & 28.6 & 30.3 \\
\midrule
M-OpenOccupancy \cite{openoccupancy}& \multirow{5}{*}{M} & 29.1 & 15.1 & 14.3 & 12.0 & 15.2 & 14.9 & 13.7 & 15.0 & 13.1 & 9.0 & 10.0 & 14.5 & 23.2 & 17.5 & 16.1 & 17.2 & 15.3 & 19.5 \\
M-CONet \cite{openoccupancy}& & 29.5 & 20.1 & 23.3 & 13.3 & 21.2 & 24.3 & 15.3 & 15.9 & 18.0 & 13.3 & 15.3 & 20.7 & 33.2 & 21.0 & 22.5 & 21.5 & 19.6 & 23.2 \\
CO-Occ \cite{coocc}& & 30.6 & 21.9 & 26.5 & 16.8 & 22.3 & 27.0 & 10.1 & 20.9 & 20.7 & 14.5 & 16.4 & 21.6 & 36.9 & 23.5 & 25.5 & 23.7 & 20.5 & 23.5 \\
OccGen \cite{occgen}& & 30.3 & 22.0 & 24.9 & 16.4 & 22.5 & 26.1 & 14.0 & 20.1 & 21.6 & 14.6 & 17.4 & 21.9 & 35.8 & 24.5 & 24.7 & 24.0 & 20.5 & 23.5 \\
\rowcolor{blue!10}
MR-Occ & & 35.5 & \textbf{27.3} & \textbf{30.5} & \textbf{22.9} & \textbf{26.6} & \textbf{30.7} & \textbf{17.3} & \textbf{28.8} & \textbf{35.4} & \textbf{21.5} & \textbf{20.7} & \textbf{26.4} & 38.1 & \textbf{26.8} & 25.9 & \textbf{26.2} & \textbf{28.6} & \textbf{30.4} \\
\bottomrule 

\end{tabular}

\endgroup 
\vspace{5pt}
\caption{3D semantic occupancy prediction results on nuScenes-Occupancy validation set. The model is trained on nuScenes-Occupancy train set and evaluated on nuScenes-Occupancy validation set. Mod.: Modality. C: Camera. L: LiDAR. M: Camera and LiDAR. const. veh.: Construction vehicle, drive. surf.: Drivable surface. The best performance is in boldface.}
\label{table:sota_v3}
\end{center}
\end{table*}

\renewcommand{\arraystretch}{1.0}

\subsection{Implementation Details}
Our framework utilizes ResNet-50 \cite{resnet} with a Feature Pyramid Network (FPN) \cite{fpn} as the camera backbone and SECOND \cite{second} as the LiDAR backbone. The input image resolution varies between datasets: for the nuScenes-Occupancy dataset, the images are set to \( 900 \times 1600 \) pixels, while for the SemanticKITTI dataset, the images are resized to \( 384 \times 1280 \) pixels to standardize the input dimensions. For LiDAR input, we apply 10 historical sweeps for the nuScenes-Occupancy dataset to improve temporal consistency, whereas for the SemanticKITTI dataset, we use the original LiDAR point clouds without any modifications.

\setlength{\tabcolsep}{2mm}  
\renewcommand{\arraystretch}{1.2}  
\begin{table}[ht!]
\begin{center}
\begin{tabular}{c|ccc|cc}  
\toprule  
Methods & PVF-Net & HVFR & OOP & IoU & mIoU\\ 
\hline\noalign{\hrule height 0.3pt}  
M-CONet &  &  &  & 29.5 & 20.1 \\
\hline\noalign{\hrule height 0pt}  
\multirow{5}{*}{\begin{tabular}[c]{@{}c@{}}MR-Occ\end{tabular}} & \checkmark &  &  & 34.4 & 25.9 \\
&  & \checkmark &  & 32.4 & 26.4 \\
&  \checkmark &  & \checkmark  & 34.5 & 26.2 \\
& \checkmark & \checkmark &  & 34.5 & 27.1 \\
& \checkmark & \checkmark & \checkmark & \textbf{35.5} & \textbf{27.3} \\
\bottomrule  
\end{tabular}
\vspace{5pt}
\caption{Ablation study on the nuScenes-Occupancy validation set. PVF-Net: Pixel to Voxel Fusion Network. HVFR: Hierarchical Voxel Feature Refinement. OOP: Occlusion-aware Occupancy Prediction.}
\label{table:ablation_main} 
\end{center}
\end{table}

The model is implemented using the MMDetection3D codebase \cite{mmdet3d}. We adopt a consistent training strategy across both datasets. Specifically, the model is trained for 15 epochs with a batch size of 1. Optimization is performed using the AdamW optimizer, with a weight decay of 0.01 and an initial learning rate of \( 3 \times 10^{-4} \). The learning rate is adjusted using a cosine learning rate scheduler with a linear warm-up phase applied during the first 500 iterations to stabilize the training process. To improve generalization, we employ data augmentation strategies inspired by BEVDet \cite{bevdet}. These include both Image Data Augmentation, such as random flipping, scaling, and photometric distortion, and BEV Data Augmentation, which consists of random rotation, scaling, and translation within the Bird’s-Eye View space. These augmentations ensure that the model can robustly handle variations in both image and LiDAR inputs during training.

All experiments are conducted on a system equipped with four NVIDIA RTX 3090 GPUs and an Intel Xeon Silver 4210R CPU. The total training time is approximately 48 hours for the nuScenes-Occupancy dataset and 16 hours for the SemanticKITTI dataset.

The nuScenes-Occupancy dataset builds upon the publicly available nuScenes dataset \cite{nuscenes}, which can be accessed at \url{https://www.nuscenes.org/}. The semantic occupancy annotations are provided by the OpenOccupancy project and are available at \url{https://github.com/JeffWang987/OpenOccupancy}. Similarly, the SemanticKITTI dataset is an extension of the KITTI Odometry Benchmark \cite{kittiodometry}, with voxel-wise semantic occupancy annotations publicly accessible at \url{http://www.semantic-kitti.org/}.

\renewcommand{\arraystretch}{1.2}  
\setlength{\tabcolsep}{7mm}  

\begin{table}[t]
\begin{center}
\setlength{\arrayrulewidth}{0.2pt} 
\begin{tabular}{c|cc}
\toprule 
Methods & IoU & mIoU\\ 
\hline\noalign{\hrule height 0.3pt} 
M-CONet & 29.5 & 20.1 \\
M-CONet-DA & 31.9 & 23.4 \\
PVF-Net w/o LD & 33.8 & 25.2 \\
PVF-Net & \textbf{34.4} & \textbf{25.9}\\
\bottomrule  
\end{tabular}
\vspace{5pt}
\caption{Ablation study of PVF-Net components. M-CONet-DA: M-CONet with 2D-to-3D view transform replaced by Deformable Attention. LD: LiDAR features Densification.}
\label{table:ablation_aff}
\end{center}
\end{table}
\renewcommand{\arraystretch}{1.2}  
\setlength{\tabcolsep}{1mm}  

\begin{table}[t]
\begin{center}
\setlength{\arrayrulewidth}{0.5pt} 
\begin{tabular}{c|ccc|cc}
\toprule 
Methods & Params & GFLOPs & FPS & IoU & mIoU\\ 
\hline\noalign{\hrule height 0.3pt} 
M-OpenOccupancy & 117M & 3045 & \textbf{4.0} & 29.1 & 15.1 \\ 
M-CONet & 137M & 3089 & 2.8 & 29.5 & 20.1 \\
Co-Occ & 205M & 2028 & 2.4 & 30.6 & 21.9 \\
OccGen & 137M & - & 2.3 & 30.3 & 22.0 \\
MR-Occ & \textbf{106M} & \textbf{1334} & 3.2 & \textbf{35.5} & \textbf{27.3}\\
\bottomrule  
\end{tabular}
\vspace{5pt}
\caption{Ablation study comparing efficiency and accuracy. All models except OccGen were inferred on an NVIDIA RTX 3090 GPU.}
\label{table:ablation_efficiency}
\end{center}
\end{table}

\renewcommand{\arraystretch}{1.2}  
\setlength{\tabcolsep}{4mm}  

\begin{table}[t]
\begin{center}
\setlength{\arrayrulewidth}{0.2pt} 
\begin{tabular}{c|c|cc}
\toprule 
Methods & Modality & IoU & mIoU\\
\hline\noalign{\hrule height 0.3pt} 
M-CONet & \multirow{3}{*}{\begin{tabular}[c]{@{}c@{}}C+L\end{tabular}} & 57.6 & 22.9 \\
Co-Occ & & 57.5 & 22.4 \\
MR-Occ & & \textbf{58.4} & \textbf{25.6}\\
\bottomrule  
\end{tabular}
\vspace{5pt}
\caption{Comparison on SemanticKITTI test set. Results for M-CONet and Co-Occ were reproduced using their respective official codes.}
\label{table:ablation_kitti}
\end{center}
\end{table}

\definecolor{roadcolor}{rgb}{1.0, 0.0, 1.0}  
\definecolor{sidewalkcolor}{rgb}{0.2, 0.0, 0.2}  
\definecolor{parkingcolor}{rgb}{1.0, 0.71, 0.76}  
\definecolor{othergrndcolor}{rgb}{0.6, 0.0, 0.2}  
\definecolor{buildingcolor}{rgb}{1.0, 0.8, 0.0}  
\definecolor{carcolor}{rgb}{0.4, 0.6, 0.8}  
\definecolor{truckcolor}{rgb}{0.0, 0.0, 0.5}  
\definecolor{bicyclecolor}{rgb}{0.0, 1.0, 1.0}  
\definecolor{motorcyclecolor}{rgb}{0.1, 0.2, 0.6}  
\definecolor{othervehcolor}{rgb}{0.4, 0.3, 0.7}
\definecolor{vegetationcolor}{rgb}{0.0, 0.7, 0.0}
\definecolor{trunckcolor}{rgb}{0.6, 0.3, 0.1}  
\definecolor{terraincolor}{rgb}{0.7, 0.9, 0.5} 
\definecolor{personcolor}{rgb}{1.0, 0.0, 0.0}  
\definecolor{bicyclistcolor}{rgb}{1.0, 0.0, 1.0}  
\definecolor{motorcyclistcolor}{rgb}{0.5, 0.0, 0.4}
\definecolor{fencecolor}{rgb}{1.0, 0.6, 0.2}
\definecolor{polecolor}{rgb}{1.0, 0.95, 0.6}
\definecolor{trafficsigncolor}{rgb}{1.0, 0.0, 0.0}

\renewcommand{\arraystretch}{1.2}  
\setlength{\tabcolsep}{2mm}  

\begin{table*}[t]
\begin{center}
\setlength{\arrayrulewidth}{0.2pt} 
\begin{tabular}{c ' c c c c c c c c c c c c c c c c c c c c}
\toprule 
Method & \rotatebox{90}{\textbf{\textcolor{roadcolor}{\rule{1em}{1em}}~road}} & \rotatebox{90}{\textbf{\textcolor{sidewalkcolor}{\rule{1em}{1em}}~sidewalk}} & \rotatebox{90}{\textbf{\textcolor{parkingcolor}{\rule{1em}{1em}}~parking}} & \rotatebox{90}{\textbf{\textcolor{othergrndcolor}{\rule{1em}{1em}}~other-ground}} & \rotatebox{90}{\textbf{\textcolor{buildingcolor}{\rule{1em}{1em}}~building}} & \rotatebox{90}{\textbf{\textcolor{carcolor}{\rule{1em}{1em}}~car}} & \rotatebox{90}{\textbf{\textcolor{truckcolor}{\rule{1em}{1em}}~truck}} & \rotatebox{90}{\textbf{\textcolor{bicyclecolor}{\rule{1em}{1em}}~bicycle}} & \rotatebox{90}{\textbf{\textcolor{motorcyclecolor}{\rule{1em}{1em}}~motorcycle}} & \rotatebox{90}{\textbf{\textcolor{othervehcolor}{\rule{1em}{1em}}~other-vehicle}} & \rotatebox{90}{\textbf{\textcolor{vegetationcolor}{\rule{1em}{1em}}~vegetation}} & \rotatebox{90}{\textbf{\textcolor{trunckcolor}{\rule{1em}{1em}}~trunk}} & \rotatebox{90}{\textbf{\textcolor{terraincolor}{\rule{1em}{1em}}~terrain}} & \rotatebox{90}{\textbf{\textcolor{personcolor}{\rule{1em}{1em}}~person}} & \rotatebox{90}{\textbf{\textcolor{bicyclistcolor}{\rule{1em}{1em}}~bicyclist}} & \rotatebox{90}{\textbf{\textcolor{motorcyclistcolor}{\rule{1em}{1em}}~motorcyclist}} & \rotatebox{90}{\textbf{\textcolor{fencecolor}{\rule{1em}{1em}}~fence}} & \rotatebox{90}{\textbf{\textcolor{polecolor}{\rule{1em}{1em}}~pole}} & \rotatebox{90}{\textbf{\textcolor{trafficsigncolor}{\rule{1em}{1em}}~traffic-sign}}\\ 
\hline\noalign{\hrule height 0.3pt} 
M-CONet \cite{openoccupancy} & 59.4 & 35.2 & 21.0 & \textbf{2.0} & 39.0 & 44.6 & 26.8 & 0.5 & 3.0 & 19.6 & 41.7 & 17.0 & 42.3 & 3.4 & \textbf{2.5} & 0.0 & 15.2 & 24.5 & 12.7 \\
CO-Occ \cite{coocc} & 59.7 & 35.2 & 21.2 & 1.7 & 39.3 & 44.1 & 28.3 & 0.5 & 3.1 & 21.4 & 41.5 & 18.1 & 42.5 & 3.4 & 2.3 & 0.0 & 14.5 & 24.2 & 12.2 \\
\rowcolor{blue!10} 
MR-Occ & \textbf{62.3} & \textbf{39.7} & \textbf{26.4} & 1.0 & \textbf{40.4} & \textbf{46.6} & \textbf{31.3} & \textbf{4.2} & \textbf{6.8} & \textbf{21.5} & \textbf{43.8} & \textbf{26.3} & \textbf{47.1} & \textbf{6.5} & \textbf{2.5} & \textbf{0.0} & \textbf{19.0} & \textbf{28.4} & \textbf{17.3} \\
\bottomrule  
\end{tabular}
\vspace{5pt}
\caption{Semantic Scene Completion results on SemanticKITTI validation set. The model is trained on SemanticKITTI train set and evaluated on SemanticKITTI validation set. All experiments were conducted using LiDAR and camera inputs. The best performance in each category is highlighted in boldface.}
\label{table:sota_kitti}
\end{center}
\end{table*}

\subsection{Main Results} 
\noindent{\bf nuScenes-Occuancy dataset.}
Table \ref{table:sota_v3} presents a performance comparison on the nuScenes-Occupancy validation set. The proposed MR-Occ model achieves state-of-the-art results in Camera-only, LiDAR-only and multimodal configurations. The C-MR-Occ model, which applies HVFR and MOD to the C-CONet model, achieved 1.7\% higher IoU and 1.8\% higher mIoU compared to C-OccGen in the Camera-only configuration. This demonstrates that the proposed approach can effectively produce reliable occupancy prediction results using only camera data. In the LiDAR-only scenario, where the deformable cross-attention module within PVF-Net is not utilized, the L-MR-Occ variant outperforms L-OccGen \cite{occgen} by 3.9\% in IoU and 7.3\% in mIoU, highlighting the robustness of the proposed voxel-level fusion strategies. Within the multimodal setting, MR-Occ surpasses the baseline M-CONet \cite{openoccupancy} by 6.0\% in IoU and 7.2\% in mIoU. When compared to the state-of-the-art OccGen model, MR-Occ provides a 5.2\% improvement in IoU and a 5.3\% improvement in mIoU. These results underscore the model’s ability to effectively integrate complementary sensor inputs and enhance spatial reasoning. Notably, while existing multimodal approaches often experience more than a 1\% reduction in IoU due to sensor misalignment, MR-Occ restricts this degradation to only 0.2\%, indicating effective mitigation of alignment challenges.

\begin{figure*}[t]
    \centering
        \includegraphics[width=1.0\linewidth]{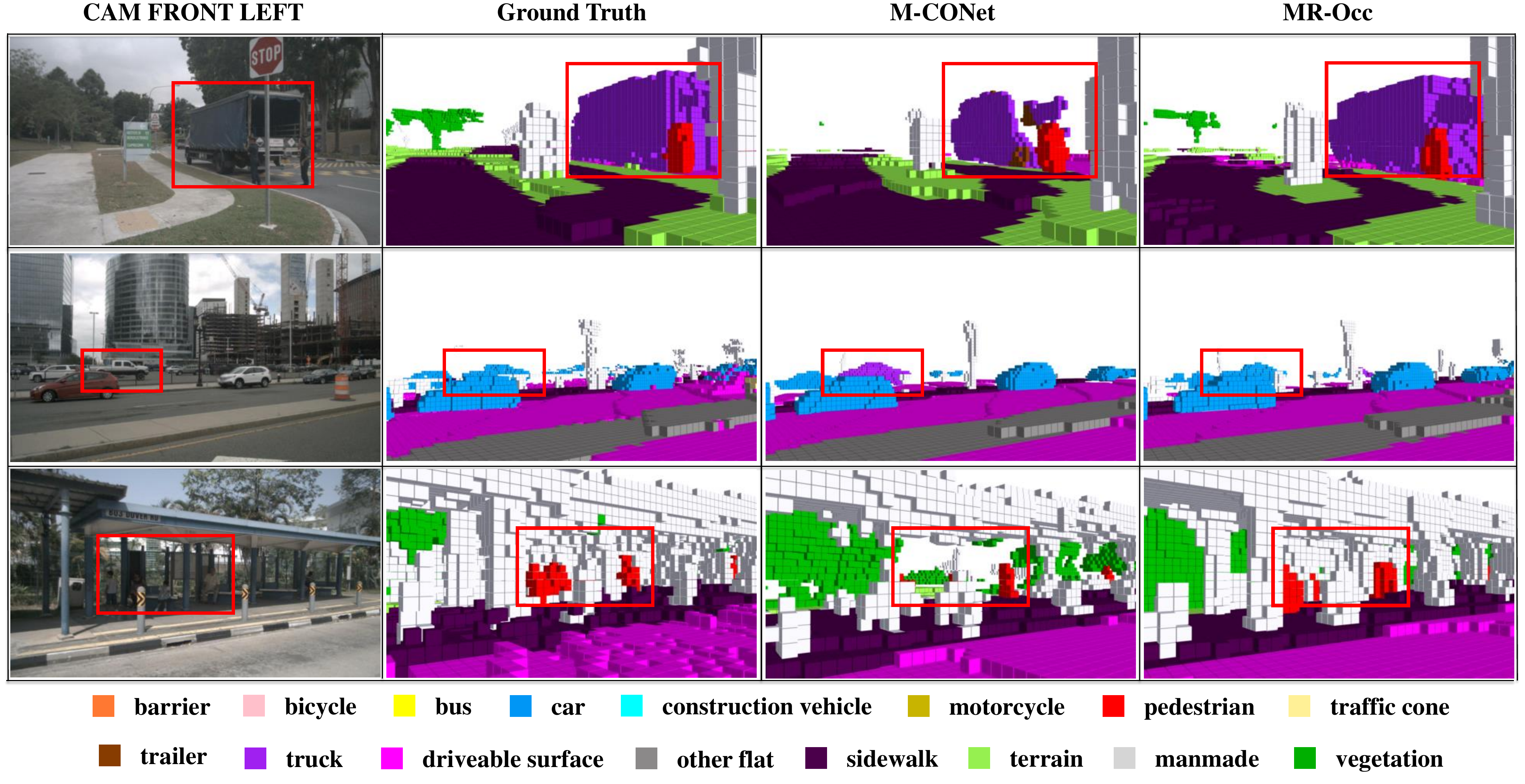}
        \caption{Qualitative results comparing MR-Occ and M-CONet predictions: The red boxes highlight areas where MR-Occ shows improved accuracy in detecting objects, particularly in object boundary regions, occlusion scenarios and small objects.}
    \label{fig:qualitative}
\end{figure*}

\renewcommand{\arraystretch}{1.2}  
\setlength{\tabcolsep}{3mm}  

\begin{table*}[t]
\begin{center}
\setlength{\arrayrulewidth}{0.2pt} 
\begin{tabular}{c|c|c|cccc|ccc|cc}
\toprule 
Methods & Metric & Standard & Sunny & Rain & Day & Night & CD\_1 & CD\_3 & CD\_5 & LBR\_4 & LBR\_16 \\
\hline\noalign{\hrule height 0.3pt} 
\multirow{2}{*}{MCONet} & IoU & 29.5 & 33.5 & 27.2 & 34.3 & 22.3 & 24.0 & 18.8 & 12.6 & 24.8 & 22.0\\ 
 & mIoU & 20.1 & 23.3 & 21.5 & 23.9 & 10.8 & 18.4 & 13.5 & 6.8 & 20.0 & 19.0\\
 \hline\noalign{\hrule height 0.1pt}
\multirow{2}{*}{MR-Occ} & IoU & 35.5 & 38.1 & 33.2 & 39.3 & 31.0 & 33.3 & 30.8 & 28.0 & 32.4 & 29.9\\ 
 & mIoU & 27.3 & 31.8 & 27.2 & 31.2 & 18.6 & 24.2 & 20.0 & 14.4 & 24.9 & 24.0\\ 
\bottomrule  
\end{tabular}
\vspace{5pt}
\caption{Robustness evaluation across various conditions. CD\_X indicates camera drop of X views, and LBR\_Y denotes LiDAR beam reduction to Y beams. Results show IoU and mIoU metrics (\%).}
\label{table:robustness} 
\end{center}
\end{table*}
\renewcommand{\arraystretch}{1.2}  
\setlength{\tabcolsep}{4mm}  

\begin{table}[t]
\begin{center}
\setlength{\arrayrulewidth}{0.2pt} 
\begin{tabular}{c|cc}
\toprule 
 & Background & Foreground\\ 
\hline\noalign{\hrule height 0.3pt} 
Coarse & 93.4 & 6.6 \\
Semi-fine & 23.5 & 76.5 \\
Fine & 3.4 & 96.6 \\
\bottomrule  
\end{tabular}
\vspace{5pt}
\caption{Region distribution analysis of RIE predictions across different resolution levels, demonstrating progressive focus on foreground regions.}
\label{table:ablation_RIE}
\end{center}
\end{table}


\subsection{Ablation studies}
\noindent{\bf Component analysis.}
Table \ref{table:ablation_main} presents an ablation study demonstrating the contribution of each component in our proposed MR-Occ on the nuScenes-Occupancy validation set. The Pixel to Voxel Fusion Network (PVF-Net), which employs deformable cross-attention guided by densified LiDAR features to fuse 2D camera features with 3D voxel features, significantly enhances performance. When applying the PVF-Net module to the baseline M-CONet, we observe significant performance improvements, with IoU increasing by 4.9\% and mIoU by 5.8\%. The Hierarchical Voxel Feature Refinement (HVFR) module, which focuses on core voxels and fuses fine-grained multimodal features, provides a significant boost when added to M-CONet, raising IoU by 2.9\% and mIoU by 6.3\%. Moreover, adding HVFR on top of the PVF-Net further increases IoU by 0.1\% and mIoU by 1.2\%. Finally, the integration of the Occlusion-aware Occupancy Prediction (OOP) module as an auxiliary task to predict occluded regions yields an additional 1.0\% improvement in IoU and 0.2\% gain in mIoU. The OOP module focused on predicting the occupancy of occluded voxels, which significantly improved IoU performance.

\noindent{\bf Effects of PVF-Net.}
Table \ref{table:ablation_aff} presents an ablation study on the nuScenes-Occupancy validation set, assessing the effectiveness of our PVF-Net components. Starting with M-CONet as the baseline, which achieves 29.5\% IoU and 20.1\% mIoU, we introduce M-CONet-DA, replacing the 2D-to-3D view transform with transformer-based feature extraction utilizing deformable attention and learnable voxel queries. This modification increases performance to 31.9\% IoU and 23.4\% mIoU. In the PVF-Net w/o LD configuration, LiDAR features are integrated into voxel queries without densification, further improving performance to 33.8\% IoU and 25.2\% mIoU, highlighting the importance of LiDAR’s spatial information. Finally, the full PVF-Net, which includes LiDAR feature densification, achieves the best results with 34.4\% IoU and 25.9\% mIoU, demonstrating that densification plays a critical role in enhancing object boundary details and overall prediction accuracy.

\noindent{\bf Efficiency and Accuracy Analysis.}
Table \ref{table:ablation_efficiency} compares the efficiency and accuracy of various models. Our proposed MR-Occ achieves state-of-the-art performance while requiring the least computational resources. Additionally, MR-Occ attains an inference speed of 3.2 FPS, the second fastest after M-OpenOccupancy. Note that our model significantly outperforms M-OpenOccupancy in accuracy, with a 6.4\% higher IoU and a 12.2\% higher mIoU, while maintaining competitive inference speed. The FPS difference stems from architectural trade-offs: while M-OpenOccupancy uses a computationally heavy Encoder but simple Occupancy Head, MR-Occ employs an efficient low-resolution Encoder but adopts M-CONet's Occupancy Head, which introduces latency through memory-intensive voxel refinement operations.

\renewcommand{\arraystretch}{1.0}
\setlength{\tabcolsep}{5mm}  

\begin{table}[t]
\begin{center}
\begin{tabular}{cc|cc}
\toprule
$\tau_{1}$ & $\tau_{2}$ & IoU & mIoU\\ \hline
0.7 & 0.4 & 35.1 & 26.5 \\
0.5 & 0.7 & 35.3 & 27.0 \\
0.4 & 0.8 & 35.1 & 27.0 \\
0.4 & 0.6 & 35.3 & 27.2 \\
0.4 & 0.7 & \textbf{35.5} & \textbf{27.3}\\
0.3 & 0.7 & 35.4 & 27.1 \\
\bottomrule
\end{tabular}
\vspace{5pt}
\caption{Ablation study on the thresholds for HVFE module using the nuScenes-Occupancy validation set. $\tau_{1}$ and $\tau_{2}$ are the thresholds for selecting the Semi-fine Resolution set \( \mathcal{S} \) and the Fine Resolution set \( \mathcal{F} \).}
\label{table:ablation_hyper}
\end{center}
\end{table}

\noindent{\bf Robustness Analysis.}
We conduct extensive experiments to evaluate MR-Occ's robustness under various challenging conditions and sensor configurations. As shown in Table \ref{table:robustness}, we analyze performance across different weather conditions (Sunny, Rain, Day, Night) and sensor degradation scenarios (camera drop and LiDAR beam reduction). MR-Occ consistently outperforms M-CONet across all test conditions. In weather scenarios, our model maintains strong performance with 38.1\% IoU in sunny conditions and demonstrates resilience in challenging scenarios like rain (33.2\% IoU) and night (31.0\% IoU). Notably, MR-Occ shows robust adaptation to sensor limitations: even with significant camera view drops (CD\_5) and LiDAR beam reductions (LBR\_16), it achieves 28.0\% IoU and 29.9\% IoU respectively, outperforming M-CONet by substantial margins (15.4\% and 7.9\% respectively). These results demonstrate that our model's hierarchical feature refinement strategy effectively maintains performance even under degraded sensor conditions.

\begin{figure*}[t]
    \centering
    \includegraphics[width=0.8\linewidth]{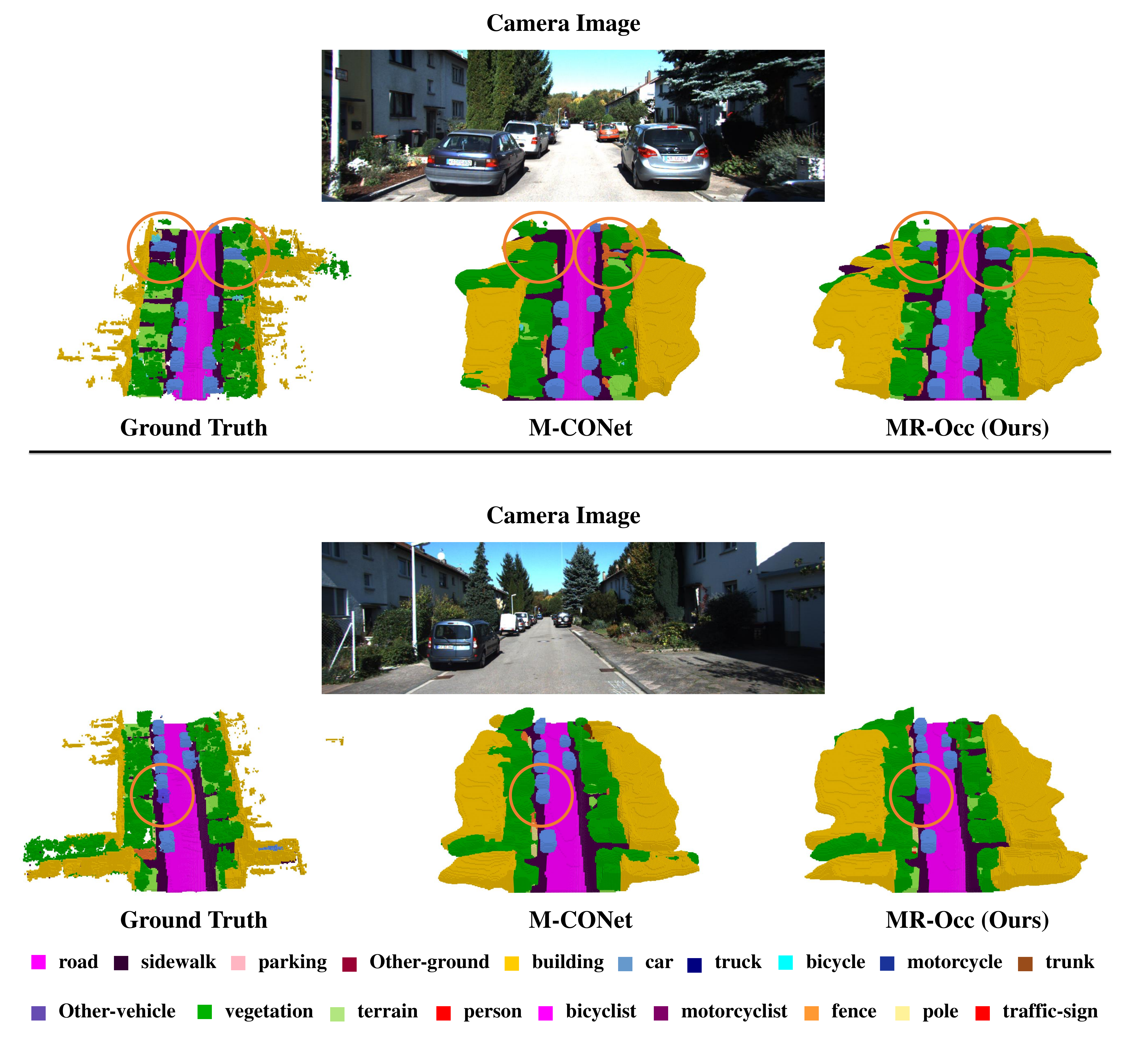}
        \caption{Qualitative comparison results on SemanticKITTI validation set. The regions highlighted by orange circles indicate areas with obvious differences.}
    \label{fig:qualitative_v2}
\end{figure*}

\noindent{\bf Resolution Importance Estimator Analysis.}
We demonstrate RIE's importance through a quantitative analysis by comparing the ratio of foreground and background regions in RIE's prediction results (Coarse voxel set ($C$), Semi-fine ($S$), and Fine Resolution ($F$)). As shown in the Table \ref{table:ablation_RIE}, the majority of $C$ corresponds to background regions (93.4\%), while $S$ and $F$ increasingly correspond to foreground regions (76.5\% and 96.6\% respectively). This indicates that the model effectively performs hierarchical predictions through RIE, enabling it to focus on foreground voxels. The progressive refinement strategy of RIE proves particularly effective, as it allows the model to allocate computational resources primarily to regions of interest. This hierarchical approach not only improves accuracy but also demonstrates computational efficiency by avoiding unnecessary processing of background regions at higher resolutions.

\noindent{\bf Hyperparameter Analysis.}
Distinguishing key voxels in the Semi-fine Resolution set $S$ and Fine Resolution set $F$ is crucial for enhancing feature representation in the Hierarchical Voxel Feature Refinement (HVFR) module. We conducted an extensive ablation study to identify the optimal thresholds $\tau_1$ and $\tau_2$ for voxel selection, as summarized in Table \ref{table:ablation_hyper}.

Different combinations of $\tau_1$ and $\tau_2$ were evaluated to assess their impact on Intersection over Union (IoU) and mean Intersection over Union (mIoU) metrics. The results demonstrate that these thresholds significantly influence the model's performance in voxel feature refinement. Among the tested configurations, setting $\tau_1 = 0.4$ and $\tau_2 = 0.7$ achieves the highest IoU of 35.5 and mIoU of 27.3. This threshold combination strikes an optimal balance by effectively capturing key voxels across different resolution scales, thus improving overall feature enhancement.

It was observed that lower values of $\tau_1$ (e.g., 0.3) or higher values of $\tau_2$ (e.g., 0.8) lead to marginally reduced performance. This reduction is likely due to the over- or under-selection of key voxels, which affects the balance between the Semi-fine and Fine Resolution sets. Despite these variations, the IoU and mIoU values remain relatively consistent across different threshold combinations, indicating the robustness of the HVFR module to small changes in threshold values.

\noindent{\bf Quantitative Results on the SemanticKITTI test set.}
We present a comparative analysis of MR-Occ's performance against reproduced results from existing models on the SemanticKITTI test set in Table \ref{table:ablation_kitti}. MR-Occ shows an improvement of 0.8\% in IoU and 2.7\% in mIoU compared to the baseline model, M-CONet.  When compared to Co-Occ, MR-Occ achieves a 0.9\% higher IoU and a 3.2\% higher mIoU. Our proposed model consistently demonstrates outstanding performance across various datasets.

Table \ref{table:sota_kitti} provides a comprehensive evaluation of semantic scene completion performance on the SemanticKITTI validation set. MR-Occ consistently outperforms existing methods across various semantic classes, underscoring its effectiveness in complex urban environments. In critical classes for autonomous driving applications, MR-Occ demonstrates significant gains. For instance, in the `car' class, our model achieves 46.6\% accuracy, surpassing M-CONet by 2.0\% and CO-Occ by 2.5\%. Similarly, MR-Occ records a 4.5\% improvement over M-CONet for the `truck' class. In challenging classes like `motorcycle' and `person,' our model outperforms M-CONet by 3.8\% and 3.1\%, respectively, demonstrating its superior capability in handling complex detection tasks. These performance gains underline MR-Occ's capability to surpass existing SOTA methods, achieving more accurate and comprehensive semantic scene completion.

\subsection{Qualitative Results}
\noindent{\bf nuScenes-Occupancy.}
Figure \ref{fig:qualitative} provides a visual comparison of 3D semantic occupancy predictions between our proposed MR-Occ and the baseline M-CONet on the nuScenes-Occupancy dataset. MR-Occ exhibits superior performance in capturing fine-grained details and accurately predicting occluded regions across various urban scenarios. In the first scene, MR-Occ accurately delineates the truck and vegetation, preserving sidewalk continuity. The second scene showcases our model’s precision in segmenting multiple cars in close proximity, maintaining clear boundaries between them. In the third scene, MR-Occ excels in predicting complex structures like the bus stop, accurately capturing glass panels and pedestrians often missed by M-CONet. These results highlight MR-Occ’s effectiveness in producing more accurate and detailed 3D semantic occupancy predictions, particularly in challenging urban environments with multiple object classes and occlusions.

\noindent{\bf SemanticKITTI.}
Figure \ref{fig:qualitative_v2} presents a visual comparison of 3D semantic occupancy predictions between MR-Occ and the M-CONet baseline on the SemanticKITTI validation set. The results clearly highlight MR-Occ's superior performance, particularly in complex urban environments.

MR-Occ effectively delineates the boundary between sidewalks and adjacent regions, providing precise predictions. It also excels in predicting the position and shape of distant and partially occluded cars, delivering reliable results even where M-CONet struggles. Additionally, MR-Occ accurately identifies object classes in complex environments where various objects are intermingled. These results demonstrate that MR-Occ effectively leverages multimodal data, ensuring robust performance across various urban environments.

\section{Conclusions}
We introduce MR-Occ, a novel and efficient camera-LiDAR fusion method for 3D semantic occupancy prediction. MR-Occ excels by utilizing a Pixel to Voxel Fusion Network, Hierarchical Voxel Feature Refinement, and a Multi-scale Occupancy Decoder, effectively addressing key challenges such as sensor misalignment and the accurate prediction of occluded regions. Our method achieves state-of-the-art performance on the nuScenes-Occupancy dataset and demonstrates highly competitive results on the SemanticKITTI dataset. This is accomplished with fewer parameters and reduced computational complexity, establishing MR-Occ as a highly efficient solution. Future work will focus on integrating temporal information to further enhance stability in dynamic environments. We hope that MR-Occ serves as a strong baseline in camera-LiDAR fusion for 3D semantic occupancy prediction, contributing valuable insights to future research in this area.


\bibliographystyle{IEEEtran}
\bibliography{IEEEtran}


\section*{Biography Section}
\vspace{-0.8cm}
\begin{IEEEbiography}[{\includegraphics[width=1in,height=1.25in,clip,keepaspectratio]{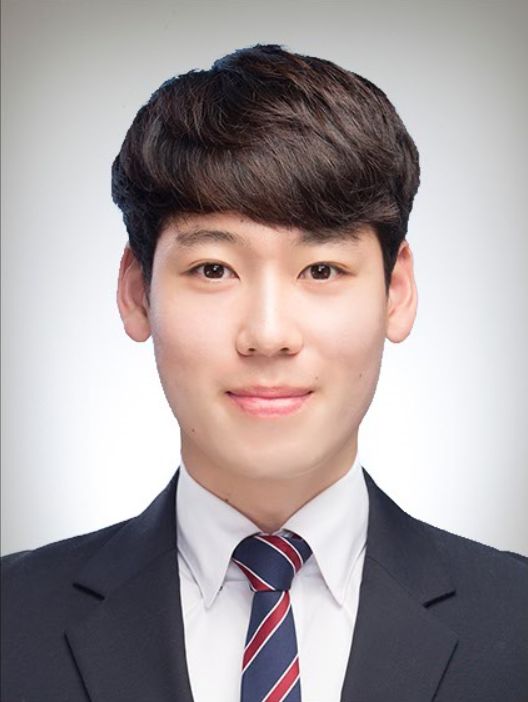}}]{Minjae Seong} received the B.S. degree in Automotive Engineering in 2020 and the M.S. degree in Artificial Intelligence in 2023, both from Hanyang University, Seoul, South Korea. He is currently pursuing the Ph.D. degree in Artificial Intelligence at Hanyang University. His research interests include deep learning-based multi-modal 3D perception, computer vision, robot perception, and autonomous driving.
\end{IEEEbiography}
\vspace{-0.8cm}
\begin{IEEEbiography}[{\includegraphics[width=1in,height=1.25in,clip,keepaspectratio]{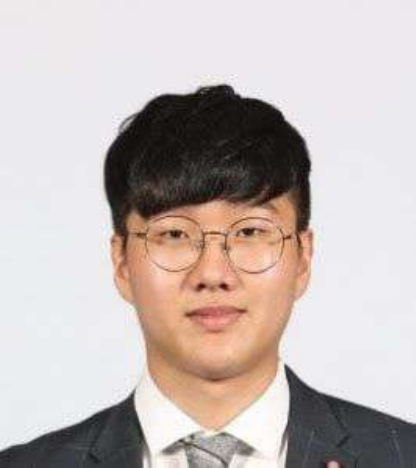}}]{Jisong Kim} received the B.S. degree in Automotive Engineering in 2020 and the M.S. degree in Electrical Engineering in 2022, both from Hanyang University, Seoul, South Korea. He is currently pursuing the Ph.D. degree in Electrical Engineering at Hanyang University. His research interests include multi-modal 3D perception, deep learning, sensor fusion, knowledge distillation, and autonomous driving.
\end{IEEEbiography}

\vspace{-0.8cm}
\begin{IEEEbiography}[{\includegraphics[width=1in,height=1.25in,clip,keepaspectratio]{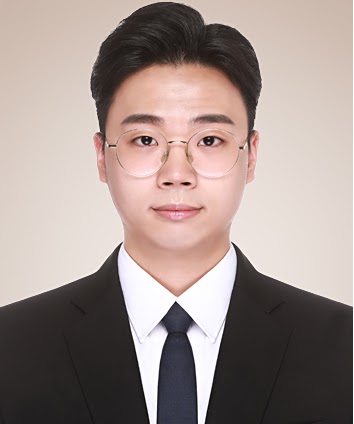}}]{Geonho Bang} received the B.S. degree in Automobile and IT Convergence in 2022 from Kookmin University, Seoul, South Korea and the M.S. degree in Artificial Intelligence from Hanyang University, Seoul, South Korea, in 2025. He is currently pursuing the Ph.D. degree in the Interdisciplinary Program in Artificial Intelligence at Seoul National University, Seoul, South Korea. His research interests include multi-modal 3D perception, sensor fusion, knowledge distillation, and autonomous driving.
\end{IEEEbiography}

\vspace{-0.8cm}
\begin{IEEEbiography}[{\includegraphics[width=1in,height=1.25in,clip,keepaspectratio]{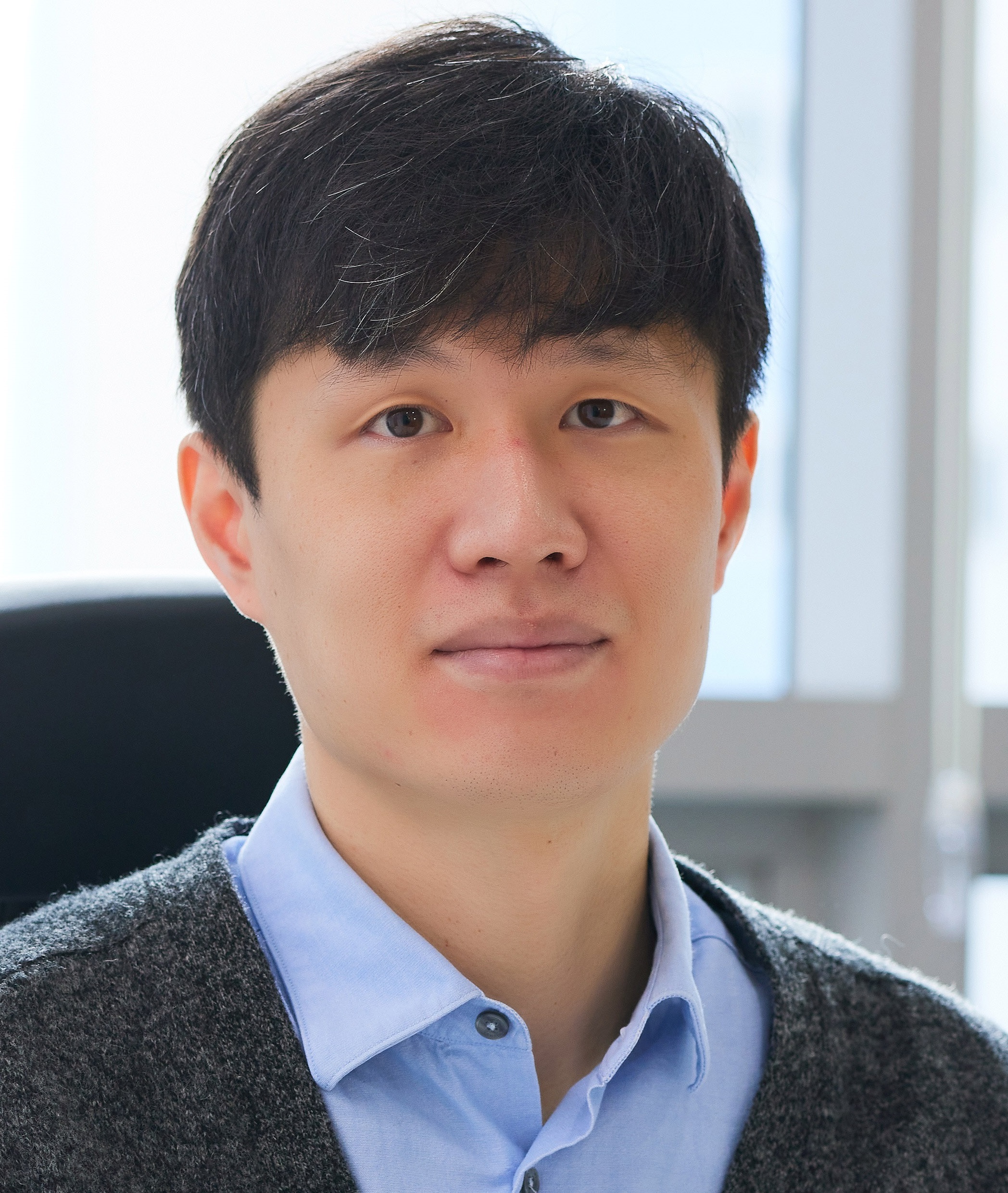}}]{Hawook Jeong} received his Ph.D. degree in Electrical and Computer Engineering from Seoul National University, Seoul, South Korea, in February 2015. He also received his M.S. degree in Electrical and Computer Engineering in 2011 and his B.S. degree in Electrical Engineering in 2009, both from Seoul National University.
Since October 2018, he has been with RideFlux, where he has served as a research scientist overseeing the development of perception and artificial intelligence algorithms for autonomous driving.
\end{IEEEbiography}

\vspace{-0.8cm}
\begin{IEEEbiography}[{\includegraphics[width=1in,height=1.25in,clip,keepaspectratio]{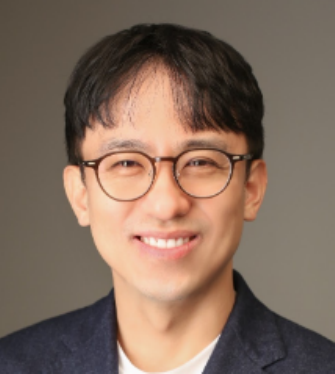}}]{Jun Won Choi} earned his B.S. and M.S. degrees from Seoul National University and his Ph.D. from the University of Illinois at Urbana-Champaign. Following his studies, he joined Qualcomm in San Diego, USA, in 2010. From 2013 to 2024, he served as a faculty member in the Department of Electrical Engineering at Hanyang University. Since 2024, he has held a faculty position in the Department of Electrical and Computer Engineering at Seoul National University. He currently serves as an Associate Editor for both IEEE Transactions on Intelligent Transportation Systems, IEEE Transactions on Vehicular Technology, International Journal of Automotive Technology. His research spans diverse areas including signal processing, machine learning, robot perception, autonomous driving, and intelligent vehicles.
\end{IEEEbiography}

\vfill

\end{document}